%% file: main.tex
\documentclass[runningheads]{llncs}
\usepackage{graphicx}

\usepackage{tikz}
\usepackage{comment}
\usepackage{amsmath,amssymb} 
\usepackage{subcaption}
\usepackage{color}
\usepackage{booktabs}
\usepackage{multirow}
\usepackage{makecell}
\newcommand{\ie}{\textit{i}.\textit{e}.}
\newcommand{\eg}{\textit{e}.\textit{g}.}

\usepackage[accsupp]{axessibility}  

\begin{document}
\pagestyle{headings}
\mainmatter
\def\ECCVSubNumber{2350}  

\title{IntereStyle: Encoding an Interest Region for Robust StyleGAN Inversion}

\titlerunning{IntereStyle}
%
\author{Seung-Jun Moon\inst{1}\index{Moon, Seung-Jun} \and
Gyeong-Moon Park\inst{2}\index{Park, Gyeong-Moon}\thanks{Corresponding Author}}
\authorrunning{S.J. Moon et al.}
%
\institute{KLleon Tech.
\email{seungjun.moon@klleon.io}\\
\and
Kyung Hee University\\
\email{gmpark@khu.ac.kr}
}
\maketitle

\input{sec/0_abstract}
\input{sec/1_introduction}
\input{sec/2_related}
\input{sec/3_method}
\input{sec/4_results}
\input{sec/6_conclusions}

\clearpage
%
%
\bibliographystyle{splncs04}
\bibliography{egbib}

\end{document}

%% file: sec/0_abstract.tex
\begin{abstract}
Recently, manipulation of real-world images has been highly elaborated along with the development of Generative Adversarial Networks (GANs) and corresponding encoders, which embed real-world images into the latent space.
However, designing encoders of GAN still remains a challenging task due to the trade-off between distortion and perception.
In this paper, we point out that the existing encoders try to lower the distortion not only on the interest region, \eg, human facial region but also on the uninterest region, \eg, background patterns and obstacles.
However, most uninterest regions in real-world images are located at out-of-distribution (OOD), which are infeasible to be ideally reconstructed by generative models.
Moreover, we empirically find that the uninterest region overlapped with the interest region can mangle the original feature of the interest region, \eg, a microphone overlapped with a facial region is inverted into the white beard.
As a result, lowering the distortion of the whole image while maintaining the perceptual quality is very challenging.
To overcome this trade-off, we propose a simple yet effective encoder training scheme, coined IntereStyle, which facilitates encoding by focusing on the interest region. IntereStyle steers the encoder to disentangle the encodings of the interest and uninterest regions.
To this end, we filter the information of the uninterest region iteratively to regulate the negative impact of the uninterest region.
We demonstrate that IntereStyle achieves both lower distortion and higher perceptual quality compared to the existing state-of-the-art encoders.
Especially, our model robustly conserves features of the original images, which shows the robust image editing and style mixing results.
We will release our code with the pre-trained model after the review.
\keywords{StyleGAN, robust GAN inversion, interest region, interest disentanglement, uninterest filter}
\end{abstract}

%% file: sec/1_introduction.tex
\section{Introduction}
\label{sec:intro}
\input{fig/teaser}
Recently, as Generative Adversarial Networks (GANs) \cite{goodfellow2014generative} have been remarkably developed, real-world image editing through latent manipulation has been prevalent \cite{shen2020interpreting,shen2020interfacegan,yang2021l2m,tov2021designing,wu2021stylespace,roich2021pivotal}. Especially, the strong disentangled property of StyleGAN \cite{karras2019style,karras2020analyzing} latent space, \ie, $W$, enables scrupulous image editing \cite{wu2021stylespace,patashnik2021styleclip}, which can change only desirable features, \eg, facial expression, while maintaining the others, \eg, identity and hairstyle.
For editing the image precisely with StyleGAN, it is required to get the suitable style latent, from which StyleGAN can reconstruct the image that has low distortion, high perceptual quality, and editability without deforming the feature of the original image.

Though StyleGAN is generally known to construct the image with high perceptual quality, the original style space $W$ is not enough to represent every real-world image with low distortion. Consequently, a vast majority of recent StyleGAN encoders, including optimization-based methods, embed images into $W+$ space \cite{abdal2019image2stylegan,richardson2021encoding,alaluf2021restyle}.
$W$ uses the identical style vector for every layer in StyleGAN, obtained by the mapping function.
On the other hand, $W+$ space provides a different style vector per layer and can even provide a random style vector in $\mathbb{R}^{512}$.
However, as the distribution of style latent is far from $W$, reconstructed images show low perceptual quality and editability \cite{tov2021designing,roich2021pivotal}.
Consequently, lowering the distortion while keeping the other factors is still challenging.

\input{fig/tradeoff}

In this paper, we claim that training to lower distortion on the entire region of the image directly is undesirable.
In most cases, images contain regions that cannot be generated due to the inherent limitation of generators.
Figure \ref{fig:teaser} shows clear examples of real-world images in the facial domain, which contain regions that are infeasible to be generated, \eg, hats, accessories, and noisy backgrounds.
Therefore, an encoder needs to concentrate on the generable region for inversion while ignoring the un-generable region (\eg, non-facial region for StyleGAN trained with FFHQ). This strategy helps the latents inverted from the generable region to be close to $W$, which show high perceptual quality and editability, as shown in Figure \ref{fig:teaser}.

Another observation is that an attempt to reconstruct the region which is not generable induces severe distortion even on the other generable regions.
For example, in Figure \ref{fig:tradeoff}, a hand overlapped with the facial region is not generable by GAN encoders. Restyle \cite{alaluf2021restyle}, which shows the lowest distortion among all encoder-based inversion models until now, tries to lower distortion on the hand too, which rather causes catastrophic distortions on the nose and chin.

In the light of these observations, it is important to distinguish the region to be reconstructed elaborately from the rest.
Here we define the term \textit{interest region}, where the model focuses on the precise reconstruction with low distortion and high perceptual quality. Practically, in most cases, the interest region is aligned with the generable region of the image.
For example, in facial image generation, the main interests are the face and hair parts of the output images, which are easier to generate than backgrounds. By focusing on the interest region, we can reduce distortion without any additional task, such as an attempt to encode latent excessively far from $W$ \cite{chen2018encoder}.

\subsubsection{Contributions}
We propose a simple yet effective method for training a StyleGAN encoder, coined IntereStyle, which steers the encoder to invert given images by focusing on the interest region.
In particular, we introduce two novel training schemes for the StyleGAN encoder: (a) Interest Disentanglement (InD) and (b) Uninterest Filter (UnF).
First, InD precludes the style latent, which includes the information on the uninterest region, from distorting the inversion result of the interest region.
Second, UnF filters the information of the uninterest region, which prevents our model from redundantly attending to the uninterest region.
UnF boosts the effect of InD by forcing the model not to focus on the uninterest region overly.
In addition, we propose a very simple yet effective scheme for determining the interest region, required only at the training stage.

We demonstrate that IntereStyle, combined with the iterative refinement \cite{alaluf2021restyle}, effectively reduces the distortion at the interest region of CelebA-HQ-test dataset.
To the best of our knowledge, IntereStyle achieves the lowest distortion among the existing state-of-the-art encoder-based StyleGAN inversion models without generator tuning.
Moreover, we qualitatively show that our model robustly preserves features of the original images even with overlapped obstacles, while other baselines fail to.
Lastly, we show the experimental results for image editing via InterFaceGAN \cite{shen2020interfacegan}, StyleCLIP \cite{patashnik2021styleclip}, and style mixing \cite{karras2019style} results, where our model shows remarkably robust outputs when input images contain significant noises, \eg, obstacles on the face.

%% file: fig/teaser.tex
\begin{figure}[t]
\centering
\includegraphics[width=0.9\columnwidth]{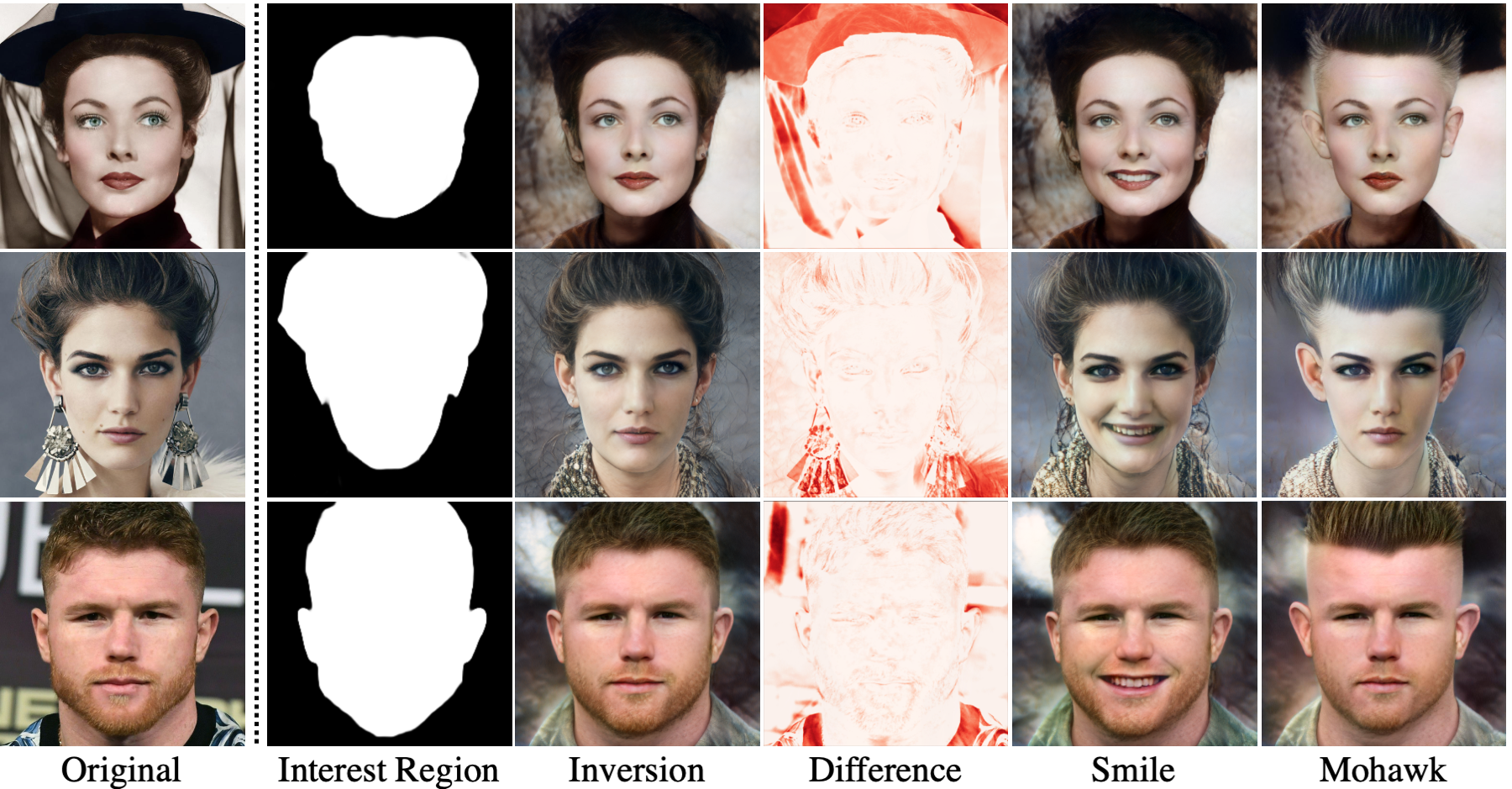}
\caption{
\textbf{Encoding of IntereStyle.}
Original images, interest region, inversion, the difference between the original images and their inversions, and the editing results (smile and Mohawk). The magnitude of the difference between the original and inversion images is colored in red. Our model successfully minimizes distortion on the interest region, even without the interest region mask for the inference. Moreover, even with low distortion, our model shows high editability.
}
\label{fig:teaser}
\end{figure}

%% file: fig/tradeoff.tex
\begin{figure}
\begin{center}
\includegraphics[width=.9\columnwidth]{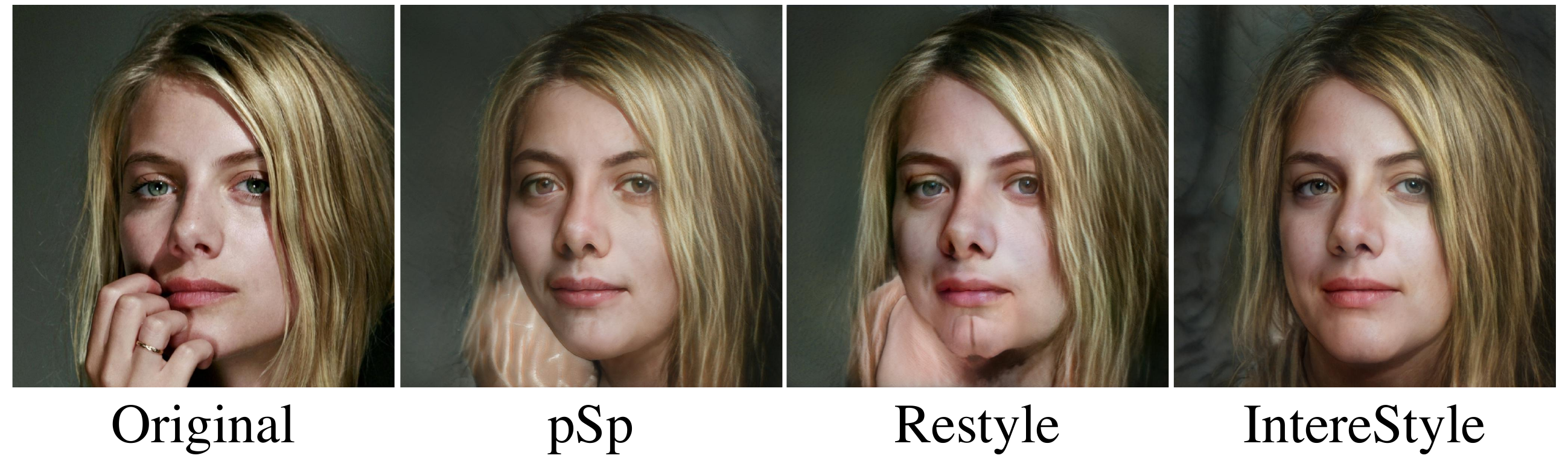}
\end{center}
\caption{
\textbf{Lowering distortion on the uninterest region.}
Inversion results of $pSp$ \cite{richardson2021encoding}, Restyle \cite{alaluf2021restyle}, and ours.
An overlapped obstacle (\ie, hand) on the facial region precludes clean inversion.
Firstly, $pSp$ shows high distortion on the eyes and generates unrealistic facial shapes on the obstacle region.
Restyle tries to reconstruct the obstacle region, but the reconstructed image shows artifacts on the nose and chin.
On the contrary, our model shows the lowest distortion among the existing models, while maintaining high perceptual quality as shown above.
}
\label{fig:tradeoff}
\end{figure}

%% file: sec/2_related.tex
\section{Related Work}
\label{sec:related}

\subsubsection{GAN Inversion}
GAN inversion aims to transform given real-world images into latent vectors from which a pre-trained GAN model can reconstruct the original image. In the early stage of GAN inversion, the majority of models rely partially \cite{zhu2016generative,bau2020semantic,bau2019seeing,zhu2020domain} or entirely \cite{creswell2018inverting,raj2019gan,abdal2019image2stylegan,abdal2020image2stylegan++,gu2020image,collins2020editing,daras2020your,saha2021loho} on the optimization steps per image.
Though the optimization-based models show high inversion qualities, these models should perform numerous optimization steps per every input image \cite{kim2021stylemapgan}, which are extremely time-consuming.
Thus, training encoders that map images into the latent space has been prevalent to invert images in the real-time domain \cite{tewari2020stylerig,zhu2020domain,richardson2021encoding,tov2021designing,alaluf2021restyle,wei2021simple}.
However, regardless of encoding methods, the existing state-of-the-art GAN inversion models focus on the whole region of images \cite{alaluf2021restyle,richardson2021encoding,tov2021designing,abdal2020image2stylegan++}, including both interest and uninterest regions.
We propose that focusing mainly on the interest region during GAN inversion improves the perceptual quality and editability of inverted images.

\input{fig/l_variation}

\subsubsection{GAN Inversion Trade-Off}
The desirable GAN inversion should consider both distortion and perceptual quality of inverted images \cite{blau2018perception,tov2021designing,roich2021pivotal}. However, due to the trade-off between two features, maintaining low distortion while enhancing the perceptual quality remains a challenging task \cite{tov2021designing,roich2021pivotal}.
Especially in StyleGAN, an inverted image from the latent far from $W$ distribution achieves lower distortion \cite{abdal2019image2stylegan,abdal2020image2stylegan++,richardson2021encoding} but shows lower perceptual quality \cite{tov2021designing} than the image from $W$ distribution.
Moreover, the latent far from $W$ distribution shows lower editability \cite{tov2021designing,roich2021pivotal}, which makes editing the inverted image harder.
Here the variance among latents for all layers can be used as an indicator of distance from $W$, where $W$ shows a zero variance due to the identical latent per layer.\footnote{Technically, we should identify whether the latents are from the mapping network of StyleGAN or not, but for simplicity, we only use the variance.}
As shown in Figure \ref{fig:variance}, the existing StyleGAN inversion models that show low distortion but suffer from low perceptual quality, \eg, $pSp$ \cite{richardson2021encoding} and Restyle \cite{alaluf2021restyle}, show relatively high variance among latents for all layers of StyleGAN.
Especially, Figure \ref{fig:l_variance_restyle} shows that Restyle gradually increases the variance of latents as the iteration refinement progresses.
In the case of $e4e$ \cite{tov2021designing}, it encodes images into latents close to $W$ but with high distortion.
In contrast to the existing methods, our model focuses on lowering distortion at the interest region, \ie, hair and face.
Since it is much easier than lowering at the uninterest region, \ie, irregular backgrounds, hats, and accessories, our model successfully achieves lower distortion than the existing models while avoiding the drop of high perceptual quality.

%% file: fig/l_variation.tex
\begin{figure}[t]
\begin{center}
\begin{subfigure}{0.45\columnwidth}
\includegraphics[width=\columnwidth]{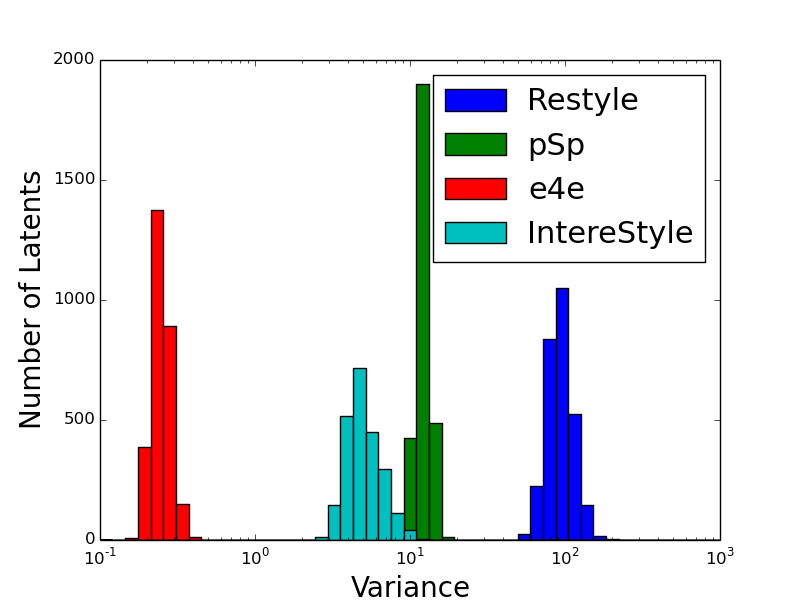}
\caption{Variance of encoders}
\label{fig:l_variance}
\end{subfigure}
\begin{subfigure}{0.45\columnwidth}
\includegraphics[width=\columnwidth]{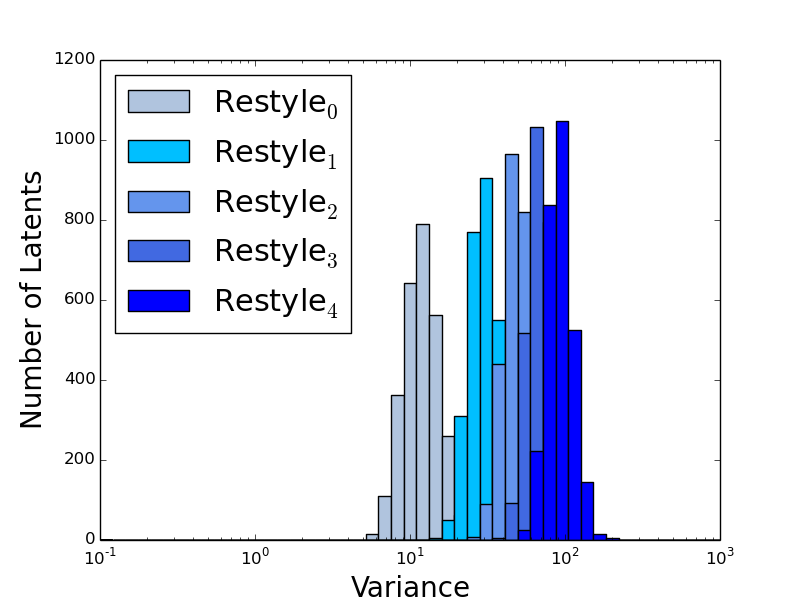}
\caption{Variance of Restyle per iteration}
\label{fig:l_variance_restyle}
\end{subfigure}
\caption{
\textbf{Variance of latents from each encoder.} We plot the variance of latents, derived from 2,800 CelebA-HQ test images with each encoder-based StyleGAN inversion model. The existing iterative refinement-based model, Restyle, shows relatively high variance among style latents per all layers. The variance of Restyle latents at iteration $i$, $\text{Restyle}_{i}$, increases along with $i$. $e4e$, which engages variation minimization loss term at the training scheme, shows relatively lower variance. We plot with a log-scale $x$-axis for better visualization.
}
\label{fig:variance}
\end{center}
\end{figure}

%% file: sec/3_method.tex
\section{Method}
\input{fig/structure}
\input{fig/dilate}
In this section, we propose a simple yet effective StyleGAN encoder training scheme named IntereStyle. We first introduce our notation and the model architecture. Then, we introduce how to determine the interest region in input images. Next, we propose two novel methods: interest disentanglement and uninterest filter in Section \ref{method:1} and Section \ref{method:2}, respectively. Finally, we describe the whole training scheme of IntereStyle in Section \ref{method:3}.

\subsection{Notation and Architecture}
\label{notation}
Our architecture is shown in Figure \ref{fig:structure}, which is based on $pSp$ \cite{richardson2021encoding} model, combined with the iterative refinement \cite{alaluf2021restyle,wei2021simple}.
At $i$-th iteration of the iterative refinement, our encoder $E$ receives a latent calculated at the previous step, $\textbf{w}_{i-1}$\footnote{When $i=1$, we utilize $\textbf{w}_{0}$ as an average latent of StyleGAN.}, together with a pair of images.
The pair consists of $(\hat{y}_{i-1}, I_{i})$, where $\hat{y}_{i-1}$ is a decoded result of $\textbf{w}_{i-1}$ via generator $G$, \ie, $\hat{y}_{i-1}=G(\textbf{w}_{i-1})$, and $I_{i}$ is a preprocessed ground truth image, $I$, by our proposed method in Section \ref{method:2}.
$E$ targets to encode the difference between $\hat{y}_{i-1}$ and $I_{i}$ into the latent, $\Delta_{i}$.
Consequently, $G$ can yield an image $\hat{y}_{i}$, which is more similar to $I_{i}$ than $\hat{y}_{i-1}$, by decoding the latent $\textbf{w}_{i}=\textbf{w}_{i-1}+\Delta_{i}$.
Our model iteratively refines the latent with a total of $N$ iterations.
Finally, we utilize a loss function $L_{image}$ for training, consisting of the weighted sum of $L_{2}$, $LPIPS$ \cite{zhang2018unreasonable}, and $L_{ID}$ \cite{richardson2021encoding}.
We explain the details of each loss in Appendix A.

\subsection{Interest Region}
\label{interestregion}
To guide the model to focus on the interest region, we need to label the interest region first.
The interest region can be designated arbitrarily according to the usage of the inversion model.
For instance, facial and hair regions for the facial domain, and the whole body for the animal domain can be set as the interest region.
For labeling this interest region, the off-the-shelf segmentation networks are used, which is described in Section \ref{experiments}.
However, directly using the segmentation masks from networks causes the distortion of facial boundaries in the generated image.
To accommodate the boundary information, we use the dilated segmentation mask containing the interest region, as shown in Figure \ref{fig:dilate}.
Without dilation, the loss term on the interest region cannot penalize the inverted face on the uninterest region.
Consequently, we dilate the mask to penalize the overflowed reconstruction of the interest region boundary.
Though the small part of the uninterest region would be included in the interest region through the dilation, we empirically find that our model still precisely generates the interest region without any distortion of boundaries.
We visually show the effect of mask dilation at the ablation study in Section \ref{qualitative}.

\subsection{Interest Disentanglement}
\label{method:1}
To enforce the model to invert the interest region into the latent space precisely, we should train the model to concentrate on the interest region regardless of the uninterest region. However, due to the spatial-agnostic feature of Adaptive Instance Normalization (AdaIN) \cite{park2019semantic}, inverted style latents considering the uninterest region may deteriorate the inversion quality of the interest region.
To prevent the encoded style of the uninterest region from deforming the inverted result of the interest region, the inversion of each region should be \textit{disentangled}.

As $E$ encodes the difference of the input pair of images in ideal, the latent $\Delta$ obtained by encoding the pair of images that only differ on the uninterest region does not contain the information of the interest region.
In other words, the decoding results from the latents $\textbf{w}$ and $\textbf{w}+\Delta$ should be the same on the interest region.
Motivated by the above, we propose a simple method named \textit{Interest Disentanglement} (InD) to distinguish the inversions of the interest region and uninterest region.
We construct the pair of input images for InD as follows: the original image $I$, and the same $I$ but multiplied with the interest region mask.
Then, as shown in Figure \ref{fig:structure}, we can yield the pair of images which only differs in the uninterest region, and the corresponding latent from $E$, $\Delta_{b}$.
Ideally, the information in $\Delta_{b}$ is solely related to the uninterest region, which implies $\textbf{w}_{N}$ generates the interest region robustly even $\Delta_{b}$ is added.
Consequently, we define InD loss, $L_{InD}$ as follows;
\begin{equation}
L_{InD}:= L_{image}(I\cdot I_{mask}, \hat{y}\cdot I_{mask}),
\end{equation}
where $\hat{y}$ is the inversion result from the latent $\textbf{w}=\textbf{w}_{N}+\Delta_{b}$.
We apply Interest Disentanglement only at the training stage, which enables the inference without the interest region mask. We empirically find that IntereStyle focuses on the interest region without any prior mask given, after the training.

\subsection{Uninterest Filter}
\label{method:2}
At the early steps of the iterative refinement, $E$ focuses on reducing the distortion of the uninterest region \cite{alaluf2021restyle}. Due to the spatial-agnostic feature of AdaIN, we claim that excessively focusing on the uninterest region hinders the inversion of the interest region. We propose a method named \textit{Uninterest Filter} (UnF), to make $E$ concentrate on the interest region at every iteration consistently.
UnF eliminates details of the uninterest region, which is inherently infeasible to reconstruct.
Thus, $E$ can reduce the redundant attention on the uninterest region for the low distortion.
In detail, UnF eases calculating $\Delta_{i}$ by blurring the uninterest region of $I$ at each iteration, with a low-pass Gaussian Filter with radius $r$, $LPF_{r}$.
As shown in Figure \ref{fig:structure}, UnF \textit{gradually} reduces the radius of Gaussian filter of $LPF$ as iterations progress, with the following two reasons; First, the redundant attention on the uninterest region is considerably severe at the early stage of the iterations \cite{alaluf2021restyle}.
Consequently, we should blur the image at the early iterations heavily. Second, excessive blur results in the severe texture difference between the interest and uninterest region.
We claim that $E$ is implicitly trained to encode the difference of the whole region, which can be biased to produce the blurred region when the blurred images are consistently given.
For the realistic generation, the input at the $N$-th iteration, $I_{N}$ is deblurred, \ie, identical to $I$. We calculate the input image at the $i$-th iteration as below:

\begin{equation}
  I_{i}=\begin{cases}
    I\cdot I_{mask} + LPF_{r_{i}}(I\cdot (1-I_{mask})), & \text{$0<i<N$}\\
    I, & \text{$i=N$}.
  \end{cases}
\end{equation}

\subsection{Training IntereStyle}
\label{method:3}
At the training stage, we jointly train the model with the off-the-shelf encoder training loss \cite{richardson2021encoding} $L_{image}$ and $L_{InD}$. However, in contrast to Restyle \cite{alaluf2021restyle}, which back-propagates $N$ times per batch, we back-propagate only once after the $N$-th iteration is over. Thus, ours show relatively faster training speed compared to Restyle.
Our final training loss is defined as below:
\begin{equation}
L_{total}:= L_{image}(I\cdot I_{mask}, \hat{y}_{N}\cdot I_{mask})+\lambda L_{InD}.
\end{equation}

Our proposed methods, InD and UnF are synergistic at the training; While InD disentangles the inversion of the uninterest region, UnF forces to look at the interest region.
Though applying $L_{image}$ to the images multiplied with $I_{mask}$ inherently drives $E$ to focus on the interest region, InD is essential for robust training.
Without $L_{InD}$, we find $E$ implicitly contains information of the uninterest region into $\Delta_{i}$, which affects the inversion of the interest region by AdaIN.

%% file: fig/structure.tex
\begin{figure*}[t]
\centering
\includegraphics[width=0.9\textwidth]{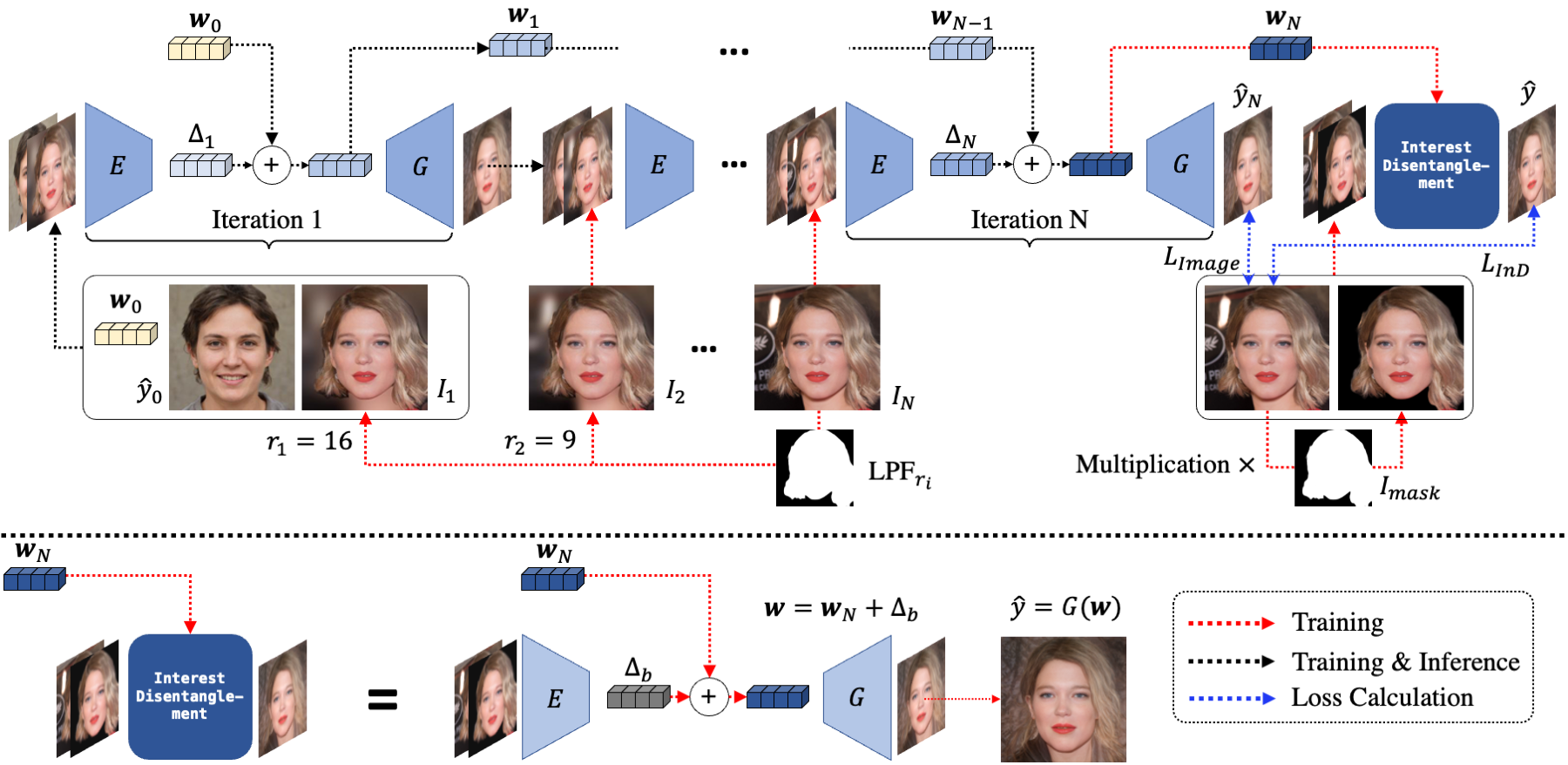}
\caption{
\textbf{Overall structure of IntereStyle.}
IntereStyle trained with total of $N$-th iteration, receives $N$ images, \ie, $I_{1},I_{2},...,I_{N}$.
$I_{i}$ is an original image passed through a low-pass filter with radius $r_{i}$, which steers the model to focus on the coarse features at the early step of iterations, and $I_{i}$ gets clearer as $i$ grows.
The Encoder targets to embed the difference of two images $I_{i}$ and $\hat{y}_{i-1}$ into the latent space, $\Delta_{i-1}$.
After the $N$-th iteration is finished, We apply interest disentanglement.
We multiply $\hat{y}_{N}$ with the $I_{mask}$, to wipe out the uninterest region.
We yield $\Delta_{b}$, by computing the difference between $I$ and $I\cdot I_{mask}$.
By adding $\Delta_{b}$ to the obtained latent $w_{N}$, we yield the output $\hat{y}$.
At the inference stage, we yield the final output $\hat{y}_{N}$, without applying interest disentanglement.
}
\label{fig:structure}
\end{figure*}

%% file: fig/dilate.tex
\begin{figure}
\begin{center}
\includegraphics[width=.9\columnwidth]{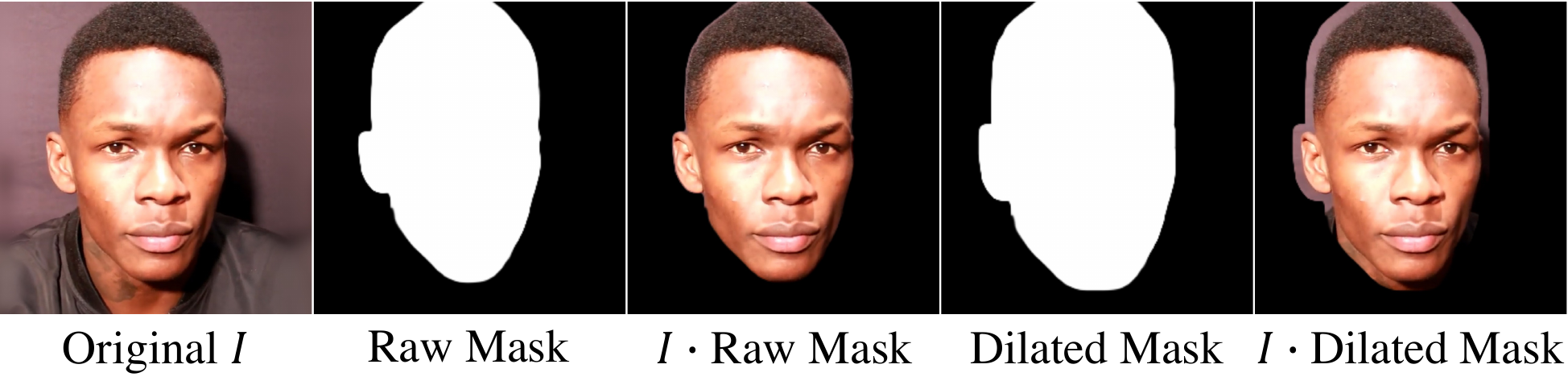}
\end{center}
\caption{
\textbf{Mask dilation.} We compare the interest region obtained by the raw mask and the dilated mask. In the raw mask, the interest region excludes the information related to the facial boundary, which occurs distortion of a facial shape. Consequently, we dilate the mask to force the interest region to include the facial boundary information, as shown in $I\cdot \text{Dilated Mask}$.}
\label{fig:dilate}
\end{figure}

%% file: sec/4_results.tex
\section{Experiments}
\label{experiments}

In this section, we briefly introduce our datasets and baselines first.
The implementation details are described in Appendix A.
Then, we compare the inversion results with baselines and ablation scenarios, both in qualitative and quantitative ways.
Next, we compare the image manipulation of our model, together with baselines.
Finally, we look into the iterative scheme of our model with Restyle.
Though we mainly show the results on the facial domain, we note that our method shows remarkable results in various domains.
We show the experimental results on the animal domain in Figure \ref{fig:main} briefly and the plenty experimental results in Appendix D.

\subsubsection{Datasets} For the facial domain, we trained the encoder using the FFHQ dataset, consisting of 70,000 human facial images. For the validation, we used the CelebA-HQ test set, consisting of 2,800 human facial images. We did not add or change any alignment or augmentation procedures compared to the existing encoder training methods \cite{richardson2021encoding,alaluf2021restyle,tov2021designing} for a fair comparison. To generate the interest and uninterest region masks, we used the pre-trained Graphonomy \cite{gong2019graphonomy} model.
For the animal domain, we used AFHQ wild dataset \cite{choi2020stargan} for training and validation, which consists of 4,730 and 500 images, respectively.
We used the pre-trained DEtection TRansformer (DE-TR; \cite{carion2020end}) for obtaining the interest region.

\subsubsection{Baselines} We compared our model with the several well-known StyleGAN encoders: IDGI \cite{zhu2020domain}, $pSp$ \cite{richardson2021encoding}, $e4e$ \cite{tov2021designing}, and Restyle \cite{alaluf2021restyle}.
Moreover, in the case of the qualitative comparison of inversion, we additionally compared it with the optimization-based model \cite{abdal2019image2stylegan,abdal2020image2stylegan++}, which is well-known for its outstanding performance.
For the baseline models, we used the pre-trained weights that are publicly available for evaluation. Please refer to Appendix C for more detailed information of each baseline.

\subsection{Inversion Evaluation}
\subsubsection{Qualitative Evaluation}
\label{qualitative}
\input{fig/main}
\input{tab/table1}

Figure \ref{fig:main} shows the inverted images of IntereStyle and two StyleGAN inversion baselines, Image2StyleGAN \cite{abdal2019image2stylegan} and Restyle \cite{alaluf2021restyle}.
In this figure, we show the entire inversion results along with the cropped images, which correspond to the areas of overlapping of the interest and uninterest regions.
Without our robust inversion schemes, mere attempts to lower distortion over the entire region often occurred severe artifacts or feature deformations.
Indeed, the baselines produced artifacts, which severely lower the perceptual quality.
In addition, they mangled features of original images in some cases.
For instance, $\text{Restyle}_{pSp}$ turned the microphone into a white beard, which does not exist in the original image.
The optimization-based inversion relatively mitigated artifacts among the baselines, but still suffered from them.
Moreover, it required more than 200 times longer inference time compared to the one of IntereStyle.
In contrast, IntereStyle showed the most robust outputs compared to all baselines, which best preserved the features of original images without artifacts.

In Figure \ref{fig:ablations}, we qualitatively showed the effectiveness of the mask dilation.
Without dilation, the model could not precisely reconstruct the original boundary of the interest region, which was denoted as the red region.
In contrast, with the mask dilation, our model minimized the boundary deformation.

\input{fig/ablations}
\input{tab/ablations}
\subsubsection{Quantitative Evaluation}

We used $L_{2}$ and LPIPS \cite{zhang2018unreasonable} losses and measured ID similarity \cite{richardson2021encoding} by utilizing the pre-trained Curricularface \cite{huang2020curricularface}, which shows the state-of-the-art performance on facial recognition.
We measured the quality on the interest region, the facial and hair regions in this paper, which need to be inverted precisely.
To this end, we multiplied the interest region mask at the calculation of ID similarity to preclude the facial recognition model from being affected by the non-facial region.
Since the facial recognition performance is dependent on the features of the non-facial region \cite{deng2019arcface}, the inverted images are prone to be identified as similar faces with the original images due to the resemblance of non-facial regions.
To compare the models solely on the facial region, we should wipe out the uninterest region.

As shown in Table \ref{tab:quantitative}, IntereStyle showed low distortion on both the interest and facial regions and preserved the identity well simultaneously.
We conclude that focusing on the interest region is indeed helpful for robust inversion.
Table \ref{tab:ablations} shows the ablation study by sequentially applying each component of our method to measure the effectiveness of the model performance.
InD reduced the negative effect of the uninterest region, which indeed lowered distortion, compared to na\"{i}vely applying $L_{image}$ on the interest region.
UnF lowered LPIPS by forcing the model to preserve features of the interest region.
Please refer to Appendix D for more detailed results of each ablation model.

\input{fig/edit}

\subsection{Editing via Latent Manipulation}
\label{edit}

Inversion of GAN is deeply related to the image manipulation on the latent space.
In this section, we compare the quality of edited images produced by various StyleGAN inversion models \cite{richardson2021encoding,tov2021designing,alaluf2021restyle}, manipulated via InterFaceGAN \cite{shen2020interfacegan} and StyleCLIP \cite{patashnik2021styleclip} methods, and style mixing \cite{karras2019style,karras2020analyzing}.
Figure \ref{fig:edit} shows the results of editing real-world images via InterFaceGAN and StyleCLIP, together with the inversion results. We changed three attributes for each method; smile, age, and pose for InterFaceGAN, and ``smile'', ``lipstick'', and ``Mohawk hairstyle'' for StyleCLIP.
Our model showed high inversion and perceptual qualities consistently among various editing scenarios, even with strong makeups or obstacles.
However, $pSp$ and $e4e$ missed important features of images, such as makeups or eye shapes. Moreover, $pSp$ produced artifacts in several editing scenarios.
In the cases of $\text{Restyle}_{pSp}$ and $\text{Restyle}_{e4e}$, they failed to robustly handle obstacles.
In the right side of Figure \ref{fig:interfacegan}, $\text{Restyle}_{pSp}$ produced severe artifacts around the mouth, while $\text{Restyle}_{e4e}$ totally changed the shape.
Moreover, the Restyle-based models showed low editability in specific cases, such as ``Mohawk''.

\input{fig/interpolation}

To attribute to the superior disentanglement feature of StyleGAN latent space \cite{shen2020interfacegan}, we can separately manipulate the coarse and fine features of images.
Following the settings from the StyleGAN \cite{karras2019style} experiment, we took styles corresponding to either coarse, middle, or fine spatial resolution, respectively, from the latent of source B, and the others were taken from the latent of source A.
Moreover, we mixed more than one hard case, \eg, obstacles on faces and extreme poses, to evaluate the robustness of our model.
As shown in Figure \ref{fig:interpolation}, our model showed higher perceptual quality on the interpolated images compared to $\text{Restyle}_{pSp}$.
$\text{Restyle}_{pSp}$ produced images with texture shift, \ie, images with cartoon-like texture, distorted facial shape, and undesirable artifacts during the style mixing. 
In contrast, our model generated stable facial outputs. Additional qualitative results are shown in Appendix D.2.

\input{fig/iter}

\subsection{Iterative Refinement of IntereStyle}

We compared the progress of iterative refinement between Restyle \cite{alaluf2021restyle} and IntereStyle in Figure \ref{fig:iter}.
Restyle reconstructed most of the coarse features within a few steps, while the variance of Restyle increased consistently as iteration progressed.
In other words, the reduction of distortion is marginal, though Restyle excessively focuses on this.
Consequently, the latent from Restyle was located far from $W$, which yields an image with low perceptual quality.
In contrast, IntereStyle concentrated on the interest region that could be generated without a broad extension from $W$.
Consequently, IntereStyle effectively reduced distortion on the interest region by iteration while maintaining high perceptual quality.

%% file: fig/main.tex
\begin{figure}[t]
\begin{center}
\includegraphics[width=.9\textwidth]{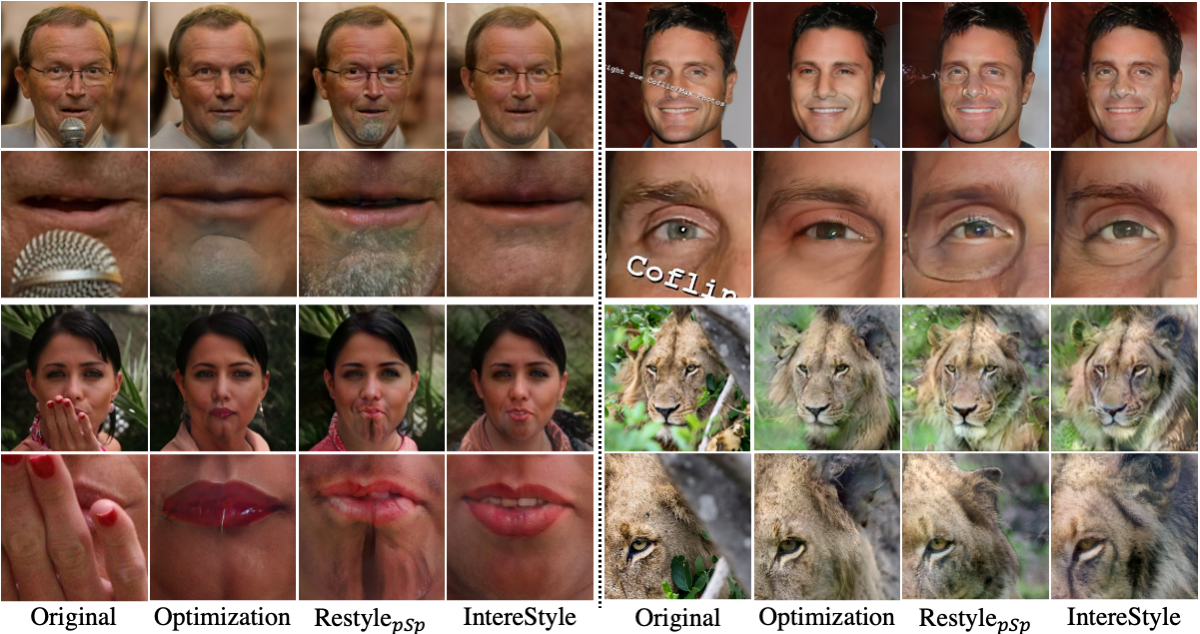}
\end{center}
\caption{
\textbf{Qualitative comparison.}
Comparison of various StyleGAN inversion methods.
IntereStyle effectively disentangled the uninterest regions (\eg, mic, letter, fingers and wood) from the interest region, which enabled robust handling of artifacts.
However, the baseline models suffered from artifacts, which significantly deformed the feature of original images. Best viewed in zoom-in.
}
\label{fig:main}
\end{figure}

%% file: tab/table1.tex
\begin{table}[h!]
\centering
\resizebox{\textwidth}{!}{
\begin{tabular}{c|cc|cc|cc|cc|cc|cc}
\toprule
Model & \multicolumn{2}{c|}{IDGI} & \multicolumn{2}{c|}{pSp} & \multicolumn{2}{c|}{e4e} & \multicolumn{2}{c|}{$\text{Restyle}_{pSp}$} & \multicolumn{2}{c|}{$\text{Restyle}_{e4e}$} & \multicolumn{2}{c}{IntereStyle} \\
Region & Interest & Face & Interest & Face & Interest & Face & Interest & Face & Interest & Face & Interest & Face \\
\midrule
\midrule
$L_{2}$ & 0.030 & 0.010 & 0.018 & 0.006 & 0.023 & 0.007 & 0.015 & 0.004 & 0.021 & 0.007 & \textbf{0.013} & \textbf{0.003}\\
LPIPS  & 0.116 & 0.053 & 0.095 & 0.046 & 0.111 & 0.051 & 0.088 & 0.040 & 0.109 & 0.054 & \textbf{0.075} & \textbf{0.036}\\
\midrule
\makecell{ID Similarity} & \multicolumn{2}{c|}{0.18} & \multicolumn{2}{c|}{0.56}& \multicolumn{2}{c|}{0.47} & \multicolumn{2}{c|}{0.65} & \multicolumn{2}{c|}{0.51} & \multicolumn{2}{c}{\textbf{0.68}}\\

\bottomrule
\end{tabular}}

\caption{\textbf{Quantitative comparison.}
We calculated each result by multiplying the mask, \ie, interest and facial masks, for the exact comparison of inversion quality of each region. IntereStyle showed the lowest $L_{2}$ and LPIPS on both the interest and facial regions among the state-of-the-art StyleGAN inversion models. Moreover, IntereStyle showed the best ID similarity among the baselines.
}
\label{tab:quantitative}
\end{table}

%% file: fig/ablations.tex
\begin{figure}
\begin{center}
\includegraphics[width=.9\columnwidth]{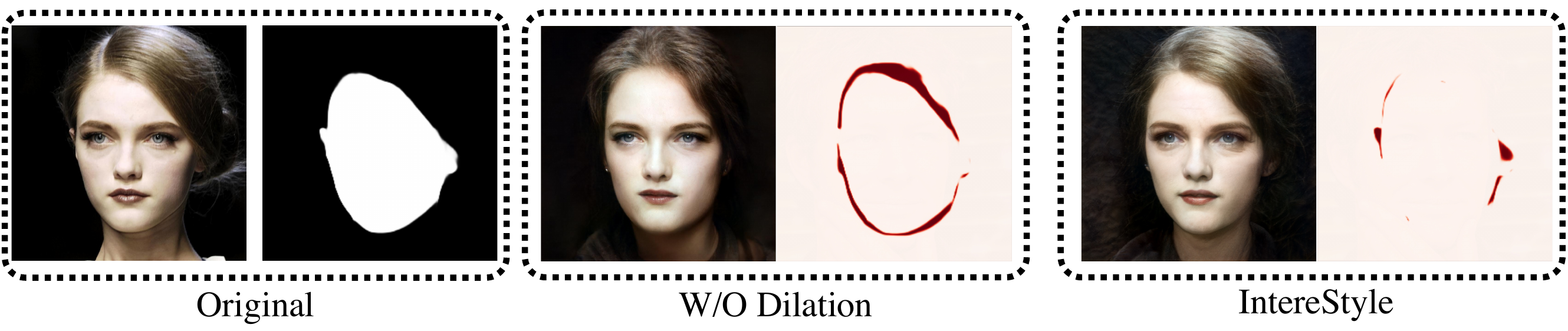}
\end{center}
\caption{
\textbf{Ablation study on dilated mask.}
Comparison of the results in the cases of using raw mask and dilated mask.
For measuring the equivalence of the facial boundaries between the original and inverted images, we used Graphonomy to obtain the facial region of each image.
We then calculated the difference of each facial region from the original one, denoted by red on the right side.
While the model without the dilated mask could not reconstruct the exact boundary, our model minimized this error effectively.
}
\label{fig:ablations}
\end{figure}

%% file: tab/ablations.tex
\begin{table}[t]
\centering
\resizebox{0.75\linewidth}{!}{
\begin{tabular}{l|cc|cc}
\toprule
Method & \multicolumn{2}{c|}{$L_{2}$} & \multicolumn{2}{c}{LPIPS} \\
& Interest & Face & Interest & Face \\
\midrule
Baseline Restyle \cite{alaluf2021restyle} & 0.015 & 0.004 & 0.088 & 0.040\\
+ $L_{image}$ on the interest region & 0.013 & 0.005 & 0.084 & 0.038 \\
+ Interest Disentanglement (InD) & \textbf{0.012} & \textbf{0.003} & 0.078 & 0.037 \\
+ Uninterest Filter (UnF) & 0.013 & \textbf{0.003} & \textbf{0.075} & \textbf{0.036} \\
\bottomrule
\end{tabular}
} 
\caption{
\textbf{Ablation study.} Performance comparison by adding each component of IntereStyle.
InD and UnF contributed to the improvement of the model, which showed better results than na\"{i}vely applying $L_{image}$ on the interest region.
Especially, InD is advantageous for reducing $L_{2}$, and UnF mainly reduces LPIPS.
}
\label{tab:ablations}
\end{table}

%% file: fig/edit.tex
\begin{figure}[t!]
\begin{center}

\begin{subfigure}{0.9\columnwidth}
\includegraphics[width=\columnwidth]{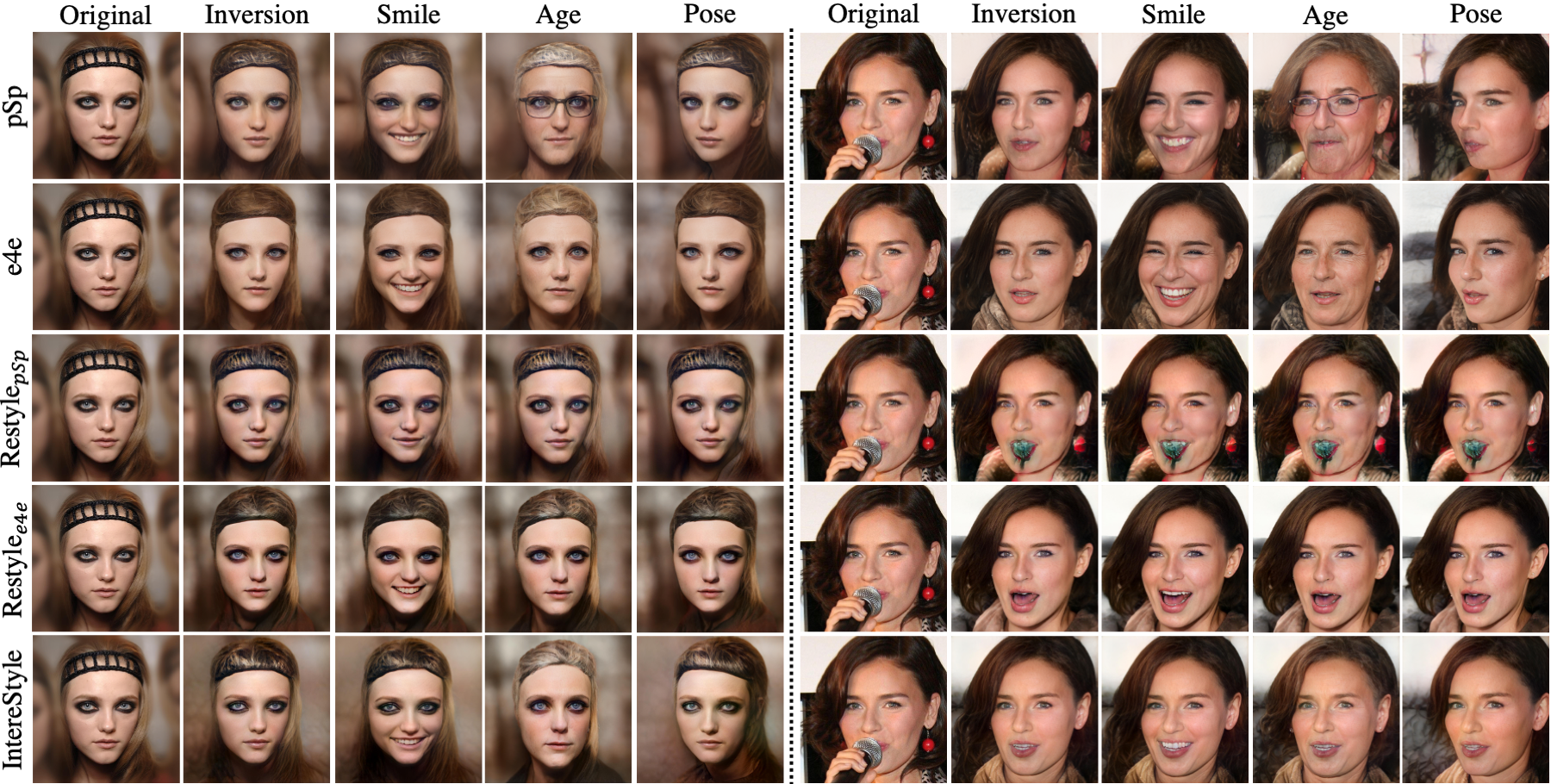}
\caption{Editing through InterFaceGAN}
\label{fig:interfacegan}
\end{subfigure}
\begin{subfigure}{0.9\columnwidth}
\includegraphics[width=\columnwidth]{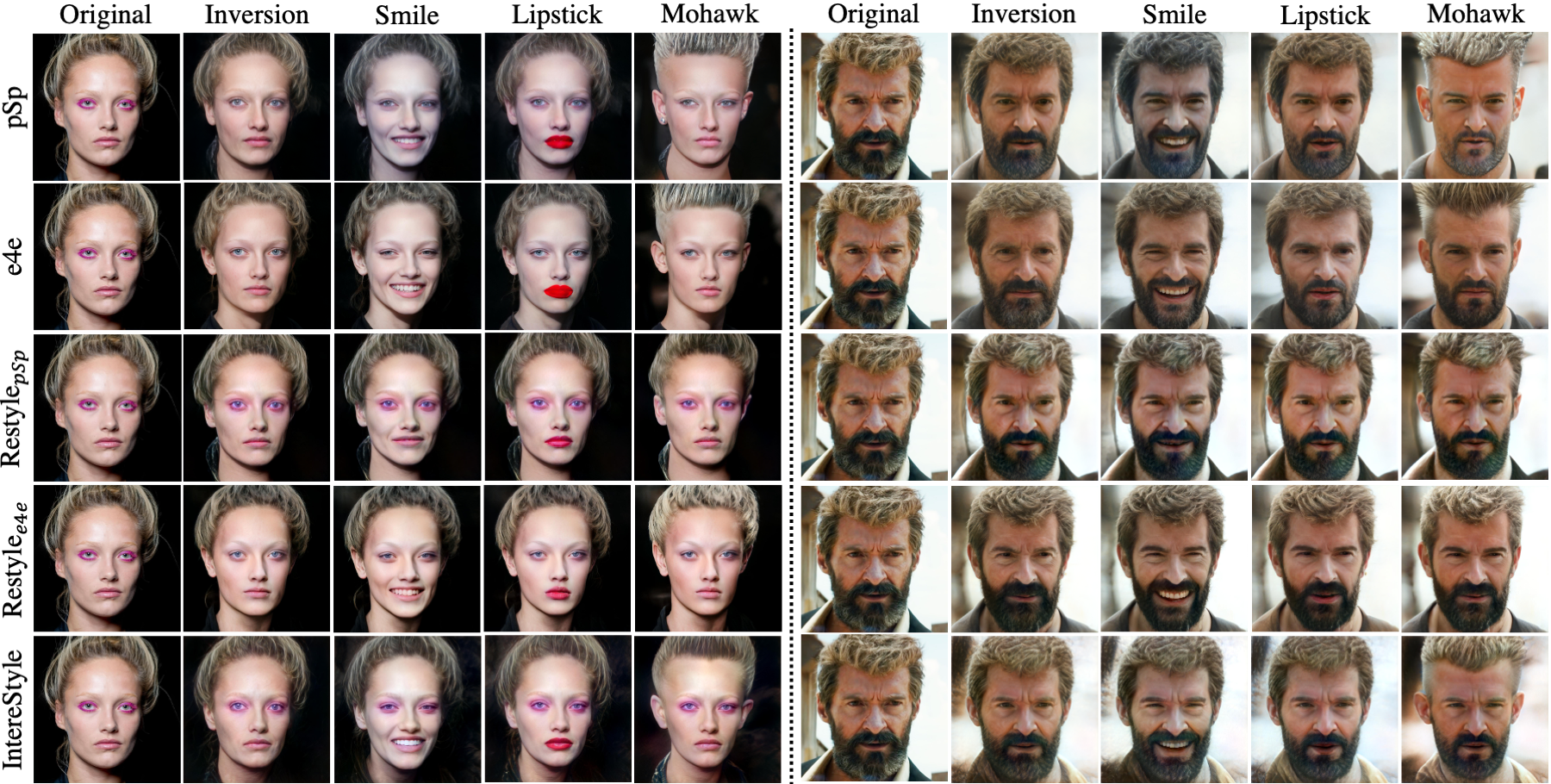}
\caption{Editing through StyleCLIP}
\label{fig:styleclip}
\end{subfigure}

\caption{
\textbf{Editing comparison.}
Edited images via (a) InterFaceGAN and (b) StyleCLIP.
Our model showed robust inversion results together with high editability consistently, while the baselines failed in various cases.
First of all, $pSp$ and $e4e$ failed to invert robustly, which ignored makeups and the detailed appearance of each image.
In the cases of $\text{Restyle}_{pSp}$ and $\text{Restyle}_{e4e}$, both were vulnerable with the overlapped obstacles. In the right image of (a), $\text{Restyle}_{pSp}$ generated severe artifacts, while $\text{Restyle}_{e4e}$ distorted the shape of mouth significantly.
Moreover, the Restyle-based models showed poor editability. In (b), $\text{Restyle}_{pSp}$ and $\text{Restyle}_{e4e}$ failed to change the hairstyle in the Mohawk case.
}
\label{fig:edit}
\end{center}
\end{figure}

%% file: fig/interpolation.tex
\begin{figure}[t]
\begin{center}
\includegraphics[width=.9\textwidth]{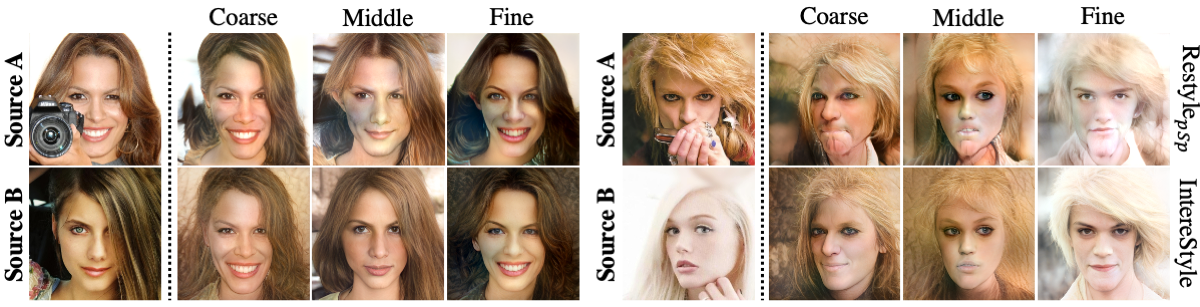}
\end{center}
\caption{
\textbf{Style mixing results.} We interpolated latents from Source A to Source B. 
We took styles corresponding to either coarse ($4^{2}-8^{2}$), middle ($16^{2}-32^{2}$), or fine ($64^{2}-1024^{2}$) resolution from source B and took the rest from source A.
IntereStyle showed the robust results on the interpolation, even with obstacles on the original source images, while $\text{Restyle}_{pSp}$ suffered from severe artifacts.
}
\label{fig:interpolation}
\end{figure}

%% file: fig/iter.tex
\begin{figure}[t]
\begin{center}
\begin{subfigure}{0.44\columnwidth}
\includegraphics[width=\columnwidth]{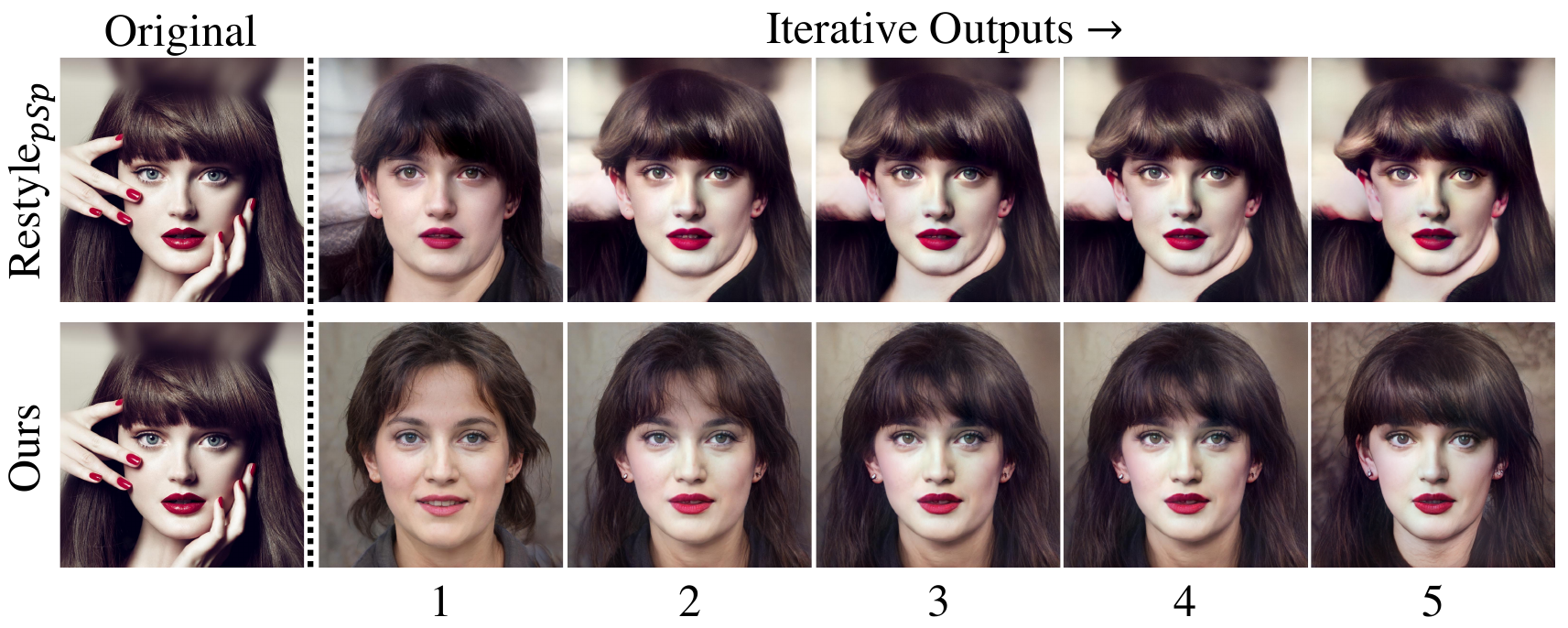}
\caption{Iterative outputs}
\label{fig:iter_1}
\end{subfigure}
\begin{subfigure}{0.44\columnwidth}
\includegraphics[width=\columnwidth]{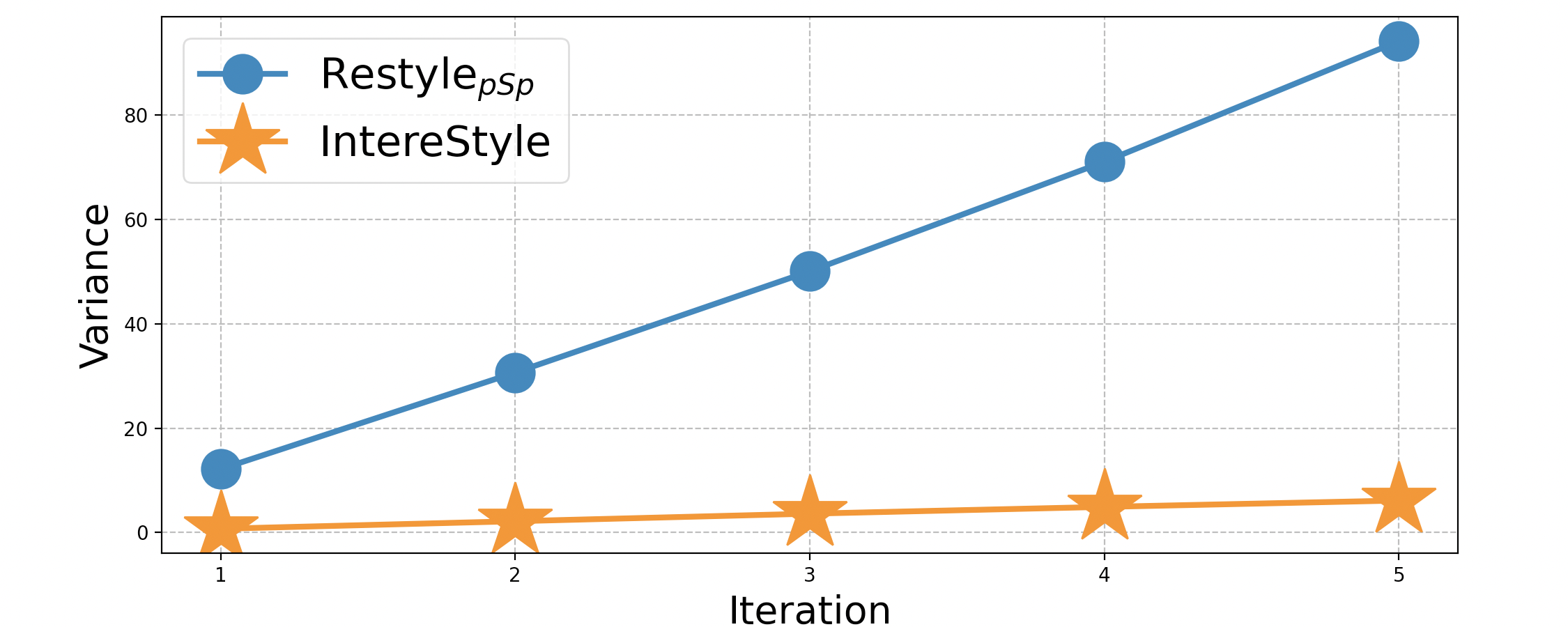}
\caption{Variance of latents}
\label{fig:iter_2}
\end{subfigure}
\caption{
\textbf{Outputs and variance at each iteration.} We compared intermediate outputs $\hat{y}_{i}$ of an example from CelebA-HQ test dataset, and the average of variances of latents among all test images at each iteration.
Restyle showed severe artifacts on the interest region and a steep variance increment at each iteration, while IntereStyle showed robust inversion and maintained low variances.
}
\label{fig:iter}
\end{center}
\end{figure}

%% file: sec/6_conclusions.tex
\section{Conclusions}
For StyleGAN inversion, focusing on the interest region is essential but under-explored yet. We found excessive attention on the uninterest region occurs the drop of perceptual quality and high distortion on the interest region. 
We proposed a simple yet effective StyleGAN encoder training scheme, coined IntereStyle, composed of Interest Disentanglement and Uninterest Filter.
We demonstrated that IntereStyle showed both low distortion and high perceptual inversion quality, and enabled various latent manipulations robustly for image editing.
We look forward to our work to be widely used in future research or the industry field, which needs a delicate inversion of the interest region for image editing.

\section{Acknowledgement}
This work was supported by the National Research Foundation of Korea(NRF) grant funded by the Korea government(MSIT) (No. 2021R1G1A1094379), and in part by the Institute of Information and Communications Technology Planning and Evaluation (IITP) grant funded by the Korea Government (MSIT) (Artificial Intelligence Innovation Hub) under Grant 2021-0-02068.

%% file: main.bbl
\begin{thebibliography}{10}
\providecommand{\url}[1]{\texttt{#1}}
\providecommand{\urlprefix}{URL }
\providecommand{\doi}[1]{https://doi.org/#1}

\bibitem{abdal2019image2stylegan}
Abdal, R., Qin, Y., Wonka, P.: Image2stylegan: How to embed images into the
  stylegan latent space? In: Proceedings of the IEEE/CVF International
  Conference on Computer Vision. pp. 4432--4441 (2019)

\bibitem{abdal2020image2stylegan++}
Abdal, R., Qin, Y., Wonka, P.: Image2stylegan++: How to edit the embedded
  images? In: Proceedings of the IEEE/CVF Conference on Computer Vision and
  Pattern Recognition. pp. 8296--8305 (2020)

\bibitem{alaluf2021restyle}
Alaluf, Y., Patashnik, O., Cohen-Or, D.: Restyle: A residual-based stylegan
  encoder via iterative refinement. In: Proceedings of the IEEE/CVF
  International Conference on Computer Vision. pp. 6711--6720 (2021)

\bibitem{bau2020semantic}
Bau, D., Strobelt, H., Peebles, W., Wulff, J., Zhou, B., Zhu, J.Y., Torralba,
  A.: Semantic photo manipulation with a generative image prior. arXiv preprint
  arXiv:2005.07727  (2020)

\bibitem{bau2019seeing}
Bau, D., Zhu, J.Y., Wulff, J., Peebles, W., Strobelt, H., Zhou, B., Torralba,
  A.: Seeing what a gan cannot generate. In: Proceedings of the IEEE/CVF
  International Conference on Computer Vision. pp. 4502--4511 (2019)

\bibitem{blau2018perception}
Blau, Y., Michaeli, T.: The perception-distortion tradeoff. In: Proceedings of
  the IEEE Conference on Computer Vision and Pattern Recognition. pp.
  6228--6237 (2018)

\bibitem{carion2020end}
Carion, N., Massa, F., Synnaeve, G., Usunier, N., Kirillov, A., Zagoruyko, S.:
  End-to-end object detection with transformers. In: European Conference on
  Computer Vision. pp. 213--229. Springer (2020)

\bibitem{chen2018encoder}
Chen, L.C., Zhu, Y., Papandreou, G., Schroff, F., Adam, H.: Encoder-decoder
  with atrous separable convolution for semantic image segmentation. In:
  Proceedings of the European conference on computer vision (ECCV). pp.
  801--818 (2018)

\bibitem{choi2020stargan}
Choi, Y., Uh, Y., Yoo, J., Ha, J.W.: Stargan v2: Diverse image synthesis for
  multiple domains. In: Proceedings of the IEEE/CVF conference on computer
  vision and pattern recognition. pp. 8188--8197 (2020)

\bibitem{collins2020editing}
Collins, E., Bala, R., Price, B., Susstrunk, S.: Editing in style: Uncovering
  the local semantics of gans. In: Proceedings of the IEEE/CVF Conference on
  Computer Vision and Pattern Recognition. pp. 5771--5780 (2020)

\bibitem{creswell2018inverting}
Creswell, A., Bharath, A.A.: Inverting the generator of a generative
  adversarial network. IEEE transactions on neural networks and learning
  systems  \textbf{30}(7),  1967--1974 (2018)

\bibitem{daras2020your}
Daras, G., Odena, A., Zhang, H., Dimakis, A.G.: Your local gan: Designing two
  dimensional local attention mechanisms for generative models. In: Proceedings
  of the IEEE/CVF Conference on Computer Vision and Pattern Recognition. pp.
  14531--14539 (2020)

\bibitem{deng2019arcface}
Deng, J., Guo, J., Xue, N., Zafeiriou, S.: Arcface: Additive angular margin
  loss for deep face recognition. In: Proceedings of the IEEE/CVF Conference on
  Computer Vision and Pattern Recognition. pp. 4690--4699 (2019)

\bibitem{gong2019graphonomy}
Gong, K., Gao, Y., Liang, X., Shen, X., Wang, M., Lin, L.: Graphonomy:
  Universal human parsing via graph transfer learning. In: Proceedings of the
  IEEE/CVF Conference on Computer Vision and Pattern Recognition. pp.
  7450--7459 (2019)

\bibitem{goodfellow2014generative}
Goodfellow, I., Pouget-Abadie, J., Mirza, M., Xu, B., Warde-Farley, D., Ozair,
  S., Courville, A., Bengio, Y.: Generative adversarial nets. Advances in
  neural information processing systems  \textbf{27} (2014)

\bibitem{gu2020image}
Gu, J., Shen, Y., Zhou, B.: Image processing using multi-code gan prior. In:
  Proceedings of the IEEE/CVF conference on computer vision and pattern
  recognition. pp. 3012--3021 (2020)

\bibitem{huang2020curricularface}
Huang, Y., Wang, Y., Tai, Y., Liu, X., Shen, P., Li, S., Li, J., Huang, F.:
  Curricularface: adaptive curriculum learning loss for deep face recognition.
  In: proceedings of the IEEE/CVF conference on computer vision and pattern
  recognition. pp. 5901--5910 (2020)

\bibitem{karras2019style}
Karras, T., Laine, S., Aila, T.: A style-based generator architecture for
  generative adversarial networks. In: Proceedings of the IEEE/CVF Conference
  on Computer Vision and Pattern Recognition. pp. 4401--4410 (2019)

\bibitem{karras2020analyzing}
Karras, T., Laine, S., Aittala, M., Hellsten, J., Lehtinen, J., Aila, T.:
  Analyzing and improving the image quality of stylegan. In: Proceedings of the
  IEEE/CVF Conference on Computer Vision and Pattern Recognition. pp.
  8110--8119 (2020)

\bibitem{kim2021stylemapgan}
Kim, H., Choi, Y., Kim, J., Yoo, S., Uh, Y.: Stylemapgan: Exploiting spatial
  dimensions of latent in gan for real-time image editing. arXiv preprint
  arXiv:2104.14754  (2021)

\bibitem{park2019semantic}
Park, T., Liu, M.Y., Wang, T.C., Zhu, J.Y.: Semantic image synthesis with
  spatially-adaptive normalization. In: Proceedings of the IEEE/CVF Conference
  on Computer Vision and Pattern Recognition. pp. 2337--2346 (2019)

\bibitem{patashnik2021styleclip}
Patashnik, O., Wu, Z., Shechtman, E., Cohen-Or, D., Lischinski, D.: Styleclip:
  Text-driven manipulation of stylegan imagery. In: Proceedings of the IEEE/CVF
  International Conference on Computer Vision. pp. 2085--2094 (2021)

\bibitem{raj2019gan}
Raj, A., Li, Y., Bresler, Y.: Gan-based projector for faster recovery with
  convergence guarantees in linear inverse problems. In: Proceedings of the
  IEEE/CVF International Conference on Computer Vision. pp. 5602--5611 (2019)

\bibitem{richardson2021encoding}
Richardson, E., Alaluf, Y., Patashnik, O., Nitzan, Y., Azar, Y., Shapiro, S.,
  Cohen-Or, D.: Encoding in style: a stylegan encoder for image-to-image
  translation. In: Proceedings of the IEEE/CVF Conference on Computer Vision
  and Pattern Recognition. pp. 2287--2296 (2021)

\bibitem{roich2021pivotal}
Roich, D., Mokady, R., Bermano, A.H., Cohen-Or, D.: Pivotal tuning for
  latent-based editing of real images. arXiv preprint arXiv:2106.05744  (2021)

\bibitem{saha2021loho}
Saha, R., Duke, B., Shkurti, F., Taylor, G.W., Aarabi, P.: Loho: Latent
  optimization of hairstyles via orthogonalization. In: Proceedings of the
  IEEE/CVF Conference on Computer Vision and Pattern Recognition. pp.
  1984--1993 (2021)

\bibitem{shen2020interpreting}
Shen, Y., Gu, J., Tang, X., Zhou, B.: Interpreting the latent space of gans for
  semantic face editing. In: Proceedings of the IEEE/CVF Conference on Computer
  Vision and Pattern Recognition. pp. 9243--9252 (2020)

\bibitem{shen2020interfacegan}
Shen, Y., Yang, C., Tang, X., Zhou, B.: Interfacegan: Interpreting the
  disentangled face representation learned by gans. IEEE transactions on
  pattern analysis and machine intelligence  (2020)

\bibitem{tewari2020stylerig}
Tewari, A., Elgharib, M., Bharaj, G., Bernard, F., Seidel, H.P., P{\'e}rez, P.,
  Zollhofer, M., Theobalt, C.: Stylerig: Rigging stylegan for 3d control over
  portrait images. In: Proceedings of the IEEE/CVF Conference on Computer
  Vision and Pattern Recognition. pp. 6142--6151 (2020)

\bibitem{tov2021designing}
Tov, O., Alaluf, Y., Nitzan, Y., Patashnik, O., Cohen-Or, D.: Designing an
  encoder for stylegan image manipulation. ACM Transactions on Graphics (TOG)
  \textbf{40}(4),  1--14 (2021)

\bibitem{wei2021simple}
Wei, T., Chen, D., Zhou, W., Liao, J., Zhang, W., Yuan, L., Hua, G., Yu, N.: A
  simple baseline for stylegan inversion. arXiv preprint arXiv:2104.07661
  (2021)

\bibitem{wu2021stylespace}
Wu, Z., Lischinski, D., Shechtman, E.: Stylespace analysis: Disentangled
  controls for stylegan image generation. In: Proceedings of the IEEE/CVF
  Conference on Computer Vision and Pattern Recognition. pp. 12863--12872
  (2021)

\bibitem{yang2021l2m}
Yang, G., Fei, N., Ding, M., Liu, G., Lu, Z., Xiang, T.: L2m-gan: Learning to
  manipulate latent space semantics for facial attribute editing. In:
  Proceedings of the IEEE/CVF Conference on Computer Vision and Pattern
  Recognition. pp. 2951--2960 (2021)

\bibitem{zhang2018unreasonable}
Zhang, R., Isola, P., Efros, A.A., Shechtman, E., Wang, O.: The unreasonable
  effectiveness of deep features as a perceptual metric. In: Proceedings of the
  IEEE conference on computer vision and pattern recognition. pp. 586--595
  (2018)

\bibitem{zhu2020domain}
Zhu, J., Shen, Y., Zhao, D., Zhou, B.: In-domain gan inversion for real image
  editing. In: European conference on computer vision. pp. 592--608. Springer
  (2020)

\bibitem{zhu2016generative}
Zhu, J.Y., Kr{\"a}henb{\"u}hl, P., Shechtman, E., Efros, A.A.: Generative
  visual manipulation on the natural image manifold. In: European conference on
  computer vision. pp. 597--613. Springer (2016)

\end{thebibliography}
